%% file: main.tex
\documentclass[10pt,twocolumn,letterpaper]{article}

\usepackage[pagenumbers]{cvpr} 

\input{preamble}

\definecolor{cvprblue}{rgb}{0.21,0.49,0.74}
\usepackage[pagebackref,breaklinks,colorlinks,allcolors=cvprblue]{hyperref}

\def\paperID{anonymous} 
\def\confName{CVPR}
\def\confYear{2026}

\title{Distilling Balanced Knowledge from a Biased Teacher}

\author{Seonghak Kim\\
{Agency for Defense Development (ADD), Republic of Korea} \\
{\tt\small seonghak35@gmail.com}
}

\begin{document}
\maketitle
\input{sec/0_abstract}    
\input{sec/1_intro}
\input{sec/2_related}
\input{sec/3_method}
\input{sec/4_experiment}

\input{sec/5_conclusion}
\section*{Acknowledgment}
This work was supported by the Agency for Defense Development Grant funded by the Korean Government.

{
    \small
    \bibliographystyle{ieeenat_fullname}
    \bibliography{main}
}

\end{document}

%% file: preamble.tex
\usepackage{tikz}
\usepackage{pgfplots}
\usepackage{multirow}
\usepackage{pifont}
\usepackage{subcaption}
\usepackage{algorithm}
\usepackage{alltt}
\usepackage{algpseudocode}
\newcommand{\cmark}{\ding{51}} 
\newcommand{\xmark}{\ding{55}} 

%% file: sec/0_abstract.tex
\begin{abstract}
Conventional knowledge distillation, designed for model compression, 
fails on long-tailed distributions 
because the teacher model tends to be biased toward head classes
and provides limited supervision for tail classes.
We propose Long-Tailed Knowledge Distillation (LTKD), 
a novel framework that reformulates the conventional objective into two components: 
a cross-group loss, capturing mismatches in prediction distributions across class groups (head, medium, and tail), 
and a within-group loss, capturing discrepancies within each group's distribution. 
This decomposition reveals the specific sources of the teacher's bias. 
To mitigate the inherited bias, LTKD introduces 
(1) a rebalanced cross-group loss that calibrates the teacher's group-level predictions 
and (2) a reweighted within-group loss that ensures equal contribution from all groups.
Extensive experiments on CIFAR-100-LT, TinyImageNet-LT, and ImageNet-LT demonstrate 
that LTKD significantly outperforms existing methods in both overall and tail-class accuracy, 
thereby showing its ability to distill balanced knowledge from a biased teacher 
for real-world applications.
\end{abstract}

%% file: sec/1_intro.tex
\section{Introduction}
\label{sec:introduction}
\begin{figure*}[t!]
  \centering
  \includegraphics[width=\linewidth]{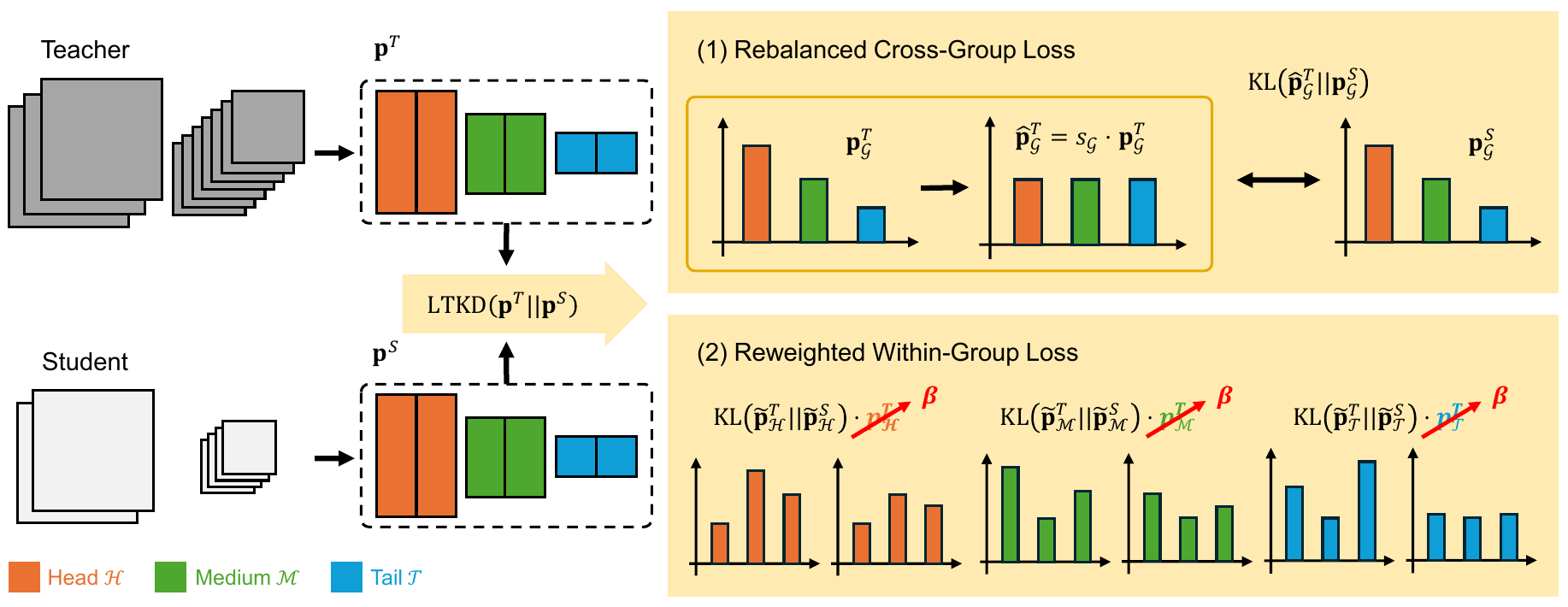}
  \caption{
  Overview of the proposed Long-Tailed Knowledge Distillation (LTKD). 
  Our method first decomposes the standard KL-based KD loss into 
  a cross-group component (capturing mismatches in aggregated group-level predictions) 
  and a within-group component (capturing internal discrepancies). 
  To correct the teacher's class bias, 
  LTKD then applies a rebalanced cross-group loss and reweighted within-group loss, 
  ensuring a balanced knowledge transfer to the student.
  }
  \label{fig:overview}
\end{figure*}

Knowledge Distillation (KD) is a well-established model compression technique 
designed to transfer knowledge from a large and powerful teacher model to a compact student model~\cite{hinton}. 
Its effectiveness has led to successful applications across diverse domains, including computer vision~\cite{li2023object, crosskd, seg_kd, yang2024vitkd}, NLP~\cite{selfkd_nlp, tinybert, sun2019patient, yang2020textbrewer}, and LLM~\cite{minillm, llmkd, yang2024survey, acharya2024survey}. 
This approach aims to retain the high accuracy of computationally expensive and large-scale models within a lightweight architecture, 
enabling deployment in resource-constrained environments~\cite{agand2024knowledge, zhao2024we, chen2024magdi}.
\begin{figure}[]
  \centering
  \includegraphics[width=\linewidth]{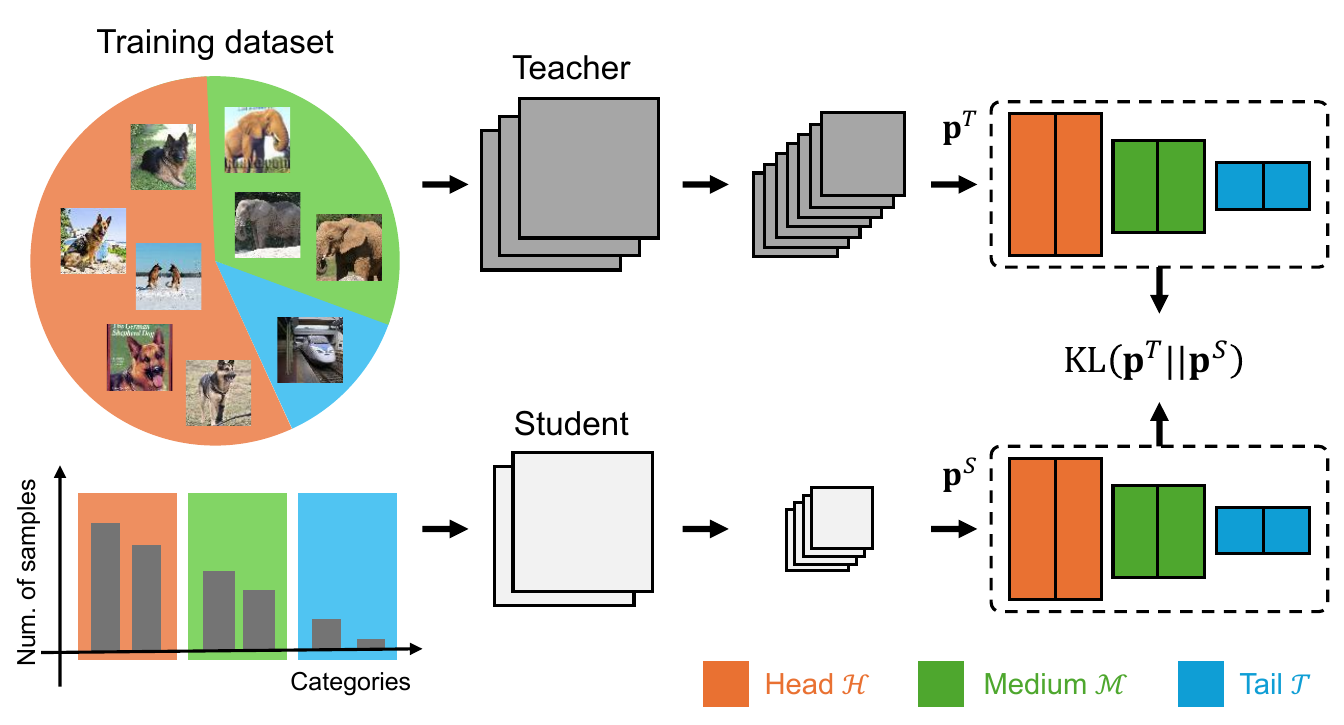}
  \caption{
  Overview of standard KD on long-tailed distributions. 
  The training data is highly imbalanced (pie chart and bar graph), with most samples belonging to head classes (orange) and few to tail classes (blue).
  This creates a biased teacher whose predictions ($\mathbf{p}^T$), visually represented by varying bar heights, are skewed toward head classes. 
  Standard KD forces the student to mimic this bias ($\mathbf{p}^S$), resulting in poor generalization on tail classes.}
  \label{fig:standard_kd}
\end{figure}
Conventional KD methods have evolved along two primary branches:
logit-based~\cite{hinton, dml, takd, dkd, dist} and feature-based~\cite{fitnet, at, rkd, crd, diffkd, freekd, review, catkd} approaches. 
Among them, logit-based methods are particularly prevalent, 
typically formulated by minimizing the Kullback–Leibler (KL) divergence 
between the teacher’s and student’s softened predictions~\cite{hinton}. 
This framework operates under a critical assumption: 
the training data is balanced~\cite{cifar, tiny}, 
allowing the teacher to offer reliable guidance across all classes. 
However, \textit{what happens when this assumption breaks down?} 
More importantly, \textit{can a teacher trained on imbalanced data still offer trustworthy supervision?}
As shown in ~\cref{fig:standard_kd},
real-world datasets often follow a long-tailed distribution~\cite{wang2022label, longtailed, longtailed2, yang2022survey}. 
When trained on such data, the teacher model becomes biased toward head classes, 
performing well on frequent classes but poorly on rare ones 
due to insufficient exposure. 
Consequently, applying standard KD is not only ineffective but can be detrimental.
The student inherits the teacher’s bias, overfitting to head-class predictions 
while receiving little meaningful guidance on tail-class examples.
This ultimately results in poor generalization. 
This highlights a critical need for new KD frameworks 
designed to distill balanced knowledge even from a biased teacher.
To address this, we propose \textit{Long-Tailed Knowledge Distillation (LTKD)}, 
a novel knowledge distillation approach tailored for class-imbalanced scenarios, 
as illustrated in \cref{fig:overview}.
Our approach reformulates the conventional KL-based KD objective 
into two components: a cross-group loss and a within-group loss. 
This decomposition enables a theoretical analysis of
how knowledge is transferred under class imbalance.
By defining appropriate class groups (\eg head, medium, and tail), 
we can elucidate the distinct contributions of these components: 
the cross-group term reflects mismatches in aggregate probability across groups, 
while the within-group term captures discrepancies within each group.
Our analysis reveals 
that both terms are distorted by the teacher's bias in distinct ways.
The cross-group term leads to an overestimation of head-class probabilities 
and an underestimation of tail-class ones. 
Meanwhile, because the within-group term is weighted by each group’s aggregate probability, 
it tends to disproportionately favor the head group while neglecting the tail group.
Building on this insight, 
we introduce two core components to counteract the teacher's bias: 
(1) a rebalanced cross-group loss that calibrates the skewed group-level predictions, 
and (2) a reweighted within-group loss that ensures equal learning focus across all groups. 
Conceptually, these adjustments offset the distortions caused by class imbalance, 
enabling the student to learn a more balanced representation.
We comprehensively validate the effectiveness of LTKD 
across a wide range of model architectures on standard long-tailed benchmarks. 
Experimental results demonstrate 
substantial improvements in both overall and, crucially, tail-class accuracy, 
consistently achieving state-of-the-art performance. 
Remarkably, in nearly all cases, 
our method surpasses the teacher’s own performance.
These results underscore the ability of LTKD
to distill balanced knowledge from a biased teacher,
marking a significant step toward deploying robust and high-performance models
in real-world and imbalanced scenarios.
In summary, our contributions are threefold:
\begin{itemize}
\item We reformulate the KL-based objective into cross-group and within-group components, enabling analysis of teacher bias under long-tailed distributions.

\item We propose rebalancing and reweighting strategies that equalize the influence of all class groups during distillation, mitigating biased supervision.

\item We achieve state-of-the-art performance on long-tailed benchmarks, improving both overall and tail-class accuracy—even surpassing the teacher in most cases.
\end{itemize}

%% file: sec/2_related.tex
\section{Related work}
\label{sec:related_work}

\subsection{Knowledge distillation on balanced datasets}
\label{subsec:kd_balanced_dataset}
Knowledge Distillation (KD) transfers rich information
—often referred to as dark knowledge~\cite{r2kd}—
from a large teacher model to a compact student, 
aiming to preserve performance while reducing computational cost~\cite{kd_survey}. 
KD methods are broadly categorized 
into feature-based~\cite{fitnet, at, rkd, crd, diffkd, freekd, review, catkd} 
and logit-based~\cite{hinton, dml, takd, dkd, dist} approaches. 
Feature-based distillation focuses on transferring internal knowledge 
by aligning hidden representations~\cite{fitnet}.
It has been extended to leverage spatial attention maps~\cite{at}, 
sample-wise relational information~\cite{rkd}, and contrastive objectives~\cite{crd}. 
ReviewKD~\cite{review} introduced a residual-style fusion of earlier feature layers to guide deeper ones,
while CAT-KD~\cite{catkd} transfers class activation maps (CAMs) 
to help the student focus on class-discriminative regions. 
Although effective, these methods typically require complex feature transformations 
and often incur higher computational overhead. 
In contrast, logit-based methods, first introduced by Hinton~\cite{hinton}, are more cost-effective 
as they aim to match the softened logits using KL divergence. 
DKD~\cite{dkd} decouples the KL divergence loss into target and non-target components 
to better reflect their distinct contributions, 
while DIST~\cite{dist} further relaxes the KL objective through correlation-based loss, 
preserving inter-class and intra-class prediction relations. 
Despite their effectiveness, 
a key limitation of existing methods is the assumption of a balanced class distribution~\cite{gou2022collaborative, catkd, diffkd, freekd}, 
which is rarely met in real-world datasets. 
A teacher trained on imbalanced data tends to be biased toward head classes, 
and this bias is directly inherited by the student. 
As a result, the student may overfit to frequent classes 
while receiving inadequate supervision for tail classes, 
leading to degraded generalization and reduced robustness.

\subsection{Knowledge distillation on LT datasets}
\label{subsec:kd_long_tailed_dataset}
A specific line of research has explored KD under long-tailed distributions~\cite{li2021self, he2021distilling, zhao2023mdcs, ju2021relational, bai2025ekdsc, bkd, lfme}, 
where the imbalance in training data leads to biased teacher predictions. 
One prominent approach is to modify the distillation objective to account for class imbalance. 
BKD~\cite{bkd} proposes a dual-loss framework 
that combines an instance-balanced classification loss 
with a class-balanced distillation loss 
to enhance performance on underrepresented classes. 
Another direction involves multi-teacher distillation. 
LFME~\cite{lfme} introduces a self-paced framework 
that aggregates knowledge from multiple expert models, 
each trained on less imbalanced subsets of the data. 
By adaptively selecting both experts and samples, 
LFME enhances generalization in long-tailed scenarios. 
However, these methods generally assume 
that the student mirrors the teacher in architecture, 
placing emphasis primarily on classification accuracy 
while largely overlooking the compression aspect.
In contrast, our work is, to the best of our knowledge, the first 
to address knowledge distillation under long-tailed distributions 
from the perspective of model compression. 
We begin by analyzing the role of KL divergence in long-tailed scenarios and demonstrate 
that the performance degradation of existing logit-based KD methods 
stems from the direct transfer of teacher bias to the student.

%% file: sec/3_method.tex
\section{Method}
\label{sec:method}
In this section, 
we begin by revisiting the KL divergence 
to diagnose the sources of teacher bias under long-tailed distributions. 
Based on this analysis, 
we propose \textit{Long-Tailed Knowledge Distillation} (LTKD), 
a framework that 
mitigates bias by rebalancing cross-group predictions 
and equalizing within-group contributions during distillation.

\subsection{Revisiting KL divergence}
\label{subsec:reformulation}
In a classification task 
where a sample $\mathbf{x}$ is categorized into one of $C$ classes, 
the predictive probability vector is represented 
as $\mathbf{p} = [p_1, p_2, \dots, p_C] \in \mathbb{R}^{C}$, 
which is obtained by applying the softmax function $\sigma (\cdot)$ 
to the logit vector $\mathbf{z} = [z_1, z_2, \dots, z_C] \in \mathbb{R}^{C}$ 
as follows:
\begin{equation}
\label{eq:pred}
    p_i=\sigma(z_i) = \frac{\exp(z_i)}{\sum^C_{j=1}\exp(z_j)},
\end{equation}
where $p_i$ and $z_i$ denote the probability and logit value 
corresponding to the $i$-th class, respectively.
Given our focus on long-tailed distributions, 
we partition the $C$ classes 
into three mutually disjoint groups, $\mathcal{G}\in\{\mathcal{H}, \mathcal{M}, \mathcal{T}\}$. 
These correspond to a head group $(\mathcal{H})$ with many samples, 
a medium group $(\mathcal{M})$, 
and a tail group $(\mathcal{T})$ with few samples.
Accordingly, the KD loss function, defined using KL divergence, can be formulated as follows:
\begin{equation}
\label{eq:kl}
\begin{aligned}
\text{KD} &= \mathrm{KL}(\mathbf{p}^T \| \mathbf{p}^S) \\
&= \sum_\mathcal{G}\sum_{i\in\mathcal{G}} p_i^T \log\left( \frac{p_i^T}{p_i^S} \right),
\end{aligned}
\end{equation}
where the teacher and student models are denoted by $T$ and $S$, respectively.
To proceed with the formulation, we introduce two new notations. 
The first is the cross-group probability distribution, 
$\mathbf{p}_\mathcal{G} = [p_\mathcal{H}, p_\mathcal{M}, p_\mathcal{T}] \in \mathbb{R}^3$, 
where each element represents the aggregated probability 
over the head, medium, and tail class groups, respectively, 
and is defined as:
\begin{equation}
\label{eq:inter_group_pred}
    p_{\mathcal{G}} = \frac{\sum_{i \in \mathcal{G}} \exp(z_i)}{\sum_{j=1}^{C} \exp(z_j)}.
\end{equation}
The second is the within-group probability distribution, 
$\tilde{\mathbf{p}}_\mathcal{G} = [\tilde{p}_{\mathcal{G}_1},\tilde{p}_{\mathcal{G}_2},\dots,\tilde{p}_{\mathcal{G}_i}]_{i \in \mathcal{G}} \in \mathbb{R}^{|\mathcal{G}|}$, 
which defines the probability distribution within each class group 
and is calculated as:
\begin{equation}
\label{eq:intra_group_pred}
    \tilde{p}_{\mathcal{G}_i} = \frac{\exp{(z_i)}}{\sum_{j \in \mathcal{G}} \exp(z_j)}.
\end{equation}
Using the relationship between $p_i$, $p_\mathcal{G}$, and $\tilde{p}_{\mathcal{G}_i}$, 
we can rewrite $p_i$ (for $i \in \mathcal{G}$) as a product of cross-group and within-group probabilities: 
$p_i = p_\mathcal{G}\cdot\tilde{p}_{\mathcal{G}_i}$. 
Based on this, \cref{eq:kl} can be reformulated as:
\begin{equation}
\label{eq:refomulated_kl}
    \begin{aligned}
    \text{KD} &= \sum_\mathcal{G}\sum_{i\in\mathcal{G}} p^T_i \log \left( \frac{p_{\mathcal{G}}^T}{p_{\mathcal{G}}^S} \right) + \sum_\mathcal{G}\sum_{i\in\mathcal{G}} p^T_i \log \left( \frac{\tilde{p}_{\mathcal{G}_i}^T}{\tilde{p}_{\mathcal{G}_i}^S} \right).
\end{aligned}
\end{equation}
Since both $p^T_\mathcal{G}$ and $p^S_\mathcal{G}$ are independent of the class index $i$, 
and $\sum_{i\in\mathcal{G}} p^T_i = p^T_\mathcal{G}$, 
the first term in \cref{eq:refomulated_kl} can be simplified as follows:
\begin{equation}
\label{eq:refomulated_first}
\begin{aligned}
\sum_\mathcal{G}\sum_{i\in\mathcal{G}} p^T_i \log \left( \frac{p_{\mathcal{G}}^T}{p_{\mathcal{G}}^S} \right) 
&= \sum_\mathcal{G}p^T_\mathcal{G}\log \left(\frac{p_{\mathcal{G}}^T}{p_{\mathcal{G}}^S}\right)  \\
&= \text{KL}\left(\mathbf{p}_\mathcal{G}^T || \mathbf{p}_\mathcal{G}^S \right),
\end{aligned}
\end{equation}
where $\mathbf{p}^T_\mathcal{G} = [p^T_\mathcal{H}, p^T_\mathcal{M}, p^T_\mathcal{T}]$ 
and $\mathbf{p}^S_\mathcal{G} = [p^S_\mathcal{H}, p^S_\mathcal{M}, p^S_\mathcal{T}]$ 
denote the cross-group probability distributions of the teacher and student, respectively. 
Analogously, the second term in \cref{eq:refomulated_kl} can also be simplified as follows:
\begin{equation}
\label{eq:refomulated_second}
\begin{aligned}
\sum_\mathcal{G}\sum_{i\in\mathcal{G}} p^T_i \log \left( \frac{\tilde{p}_{\mathcal{G}_i}^T}{\tilde{p}_{\mathcal{G}_i}^S} \right) 
&= \sum_\mathcal{G}p^T_\mathcal{G}\sum_{i\in \mathcal{G}}\tilde{p}_{\mathcal{G}_i}^T\log \left(\frac{\tilde{p}_{\mathcal{G}_i}^T}{\tilde{p}_{\mathcal{G}_i}^S}\right) \\
&= \sum_\mathcal{G}p^T_\mathcal{G}\cdot\text{KL}\left( \tilde{\mathbf{p}}^T_{\mathcal{{G}}} || \tilde{\mathbf{p}}^S_{\mathcal{{G}}} \right),
\end{aligned}
\end{equation}
where $\tilde{\mathbf{p}}^T_{\mathcal{{G}}}$ and $\tilde{\mathbf{p}}^S_{\mathcal{{G}}}$ 
denote the within-group probability distributions of the teacher and student 
for group $\mathcal{G}$, respectively.
Finally, \cref{eq:kl} can be expressed as:
\begin{equation}
\label{eq:new_kl}
\begin{aligned}
\text{KD} &= \text{KL} \left( \mathbf{p}^T_\mathcal{G} \| \mathbf{p}^S_\mathcal{G}\right) + \sum_\mathcal{G}p^T_\mathcal{G}\cdot\text{KL}\left( \tilde{\mathbf{p}}^T_\mathcal{G} \| \tilde{\mathbf{p}}^S_\mathcal{G} \right) \\
&= \text{KL} \left( \mathbf{p}^T_\mathcal{G} \| \mathbf{p}^S_\mathcal{G}\right)+p^T_\mathcal{H} \cdot \text{KL}\left( \tilde{\mathbf{p}}^T_\mathcal{H} \| \tilde{\mathbf{p}}^S_\mathcal{H} \right) \\
&+ p^T_\mathcal{M} \cdot \text{KL}\left( \tilde{\mathbf{p}}^T_\mathcal{M} \| \tilde{\mathbf{p}}^S_\mathcal{M} \right) + p^T_\mathcal{T} \cdot \text{KL}\left( \tilde{\mathbf{p}}^T_\mathcal{T} \| \tilde{\mathbf{p}}^S_\mathcal{T} \right).
\end{aligned}
\end{equation}
This formulation reveals that the overall KD loss consists of two key components: 
(1) cross-group loss, 
which aligns the aggregate probability distributions across the class groups (head, medium, and tail); 
(2) weighted sum of within-group losses, 
which aligns the probability distributions within each group. 
Notably, the weight assigned to each within-group loss 
corresponds to the teacher's cross-group probability $p^T_\mathcal{G}$.
This decomposition 
enables analysis of the distinct contributions from cross- and within-group learning, 
revealing the inherent limitations of existing KD in long-tailed settings.

\begin{figure}[t!]
  \centering
  \includegraphics[width=\linewidth]{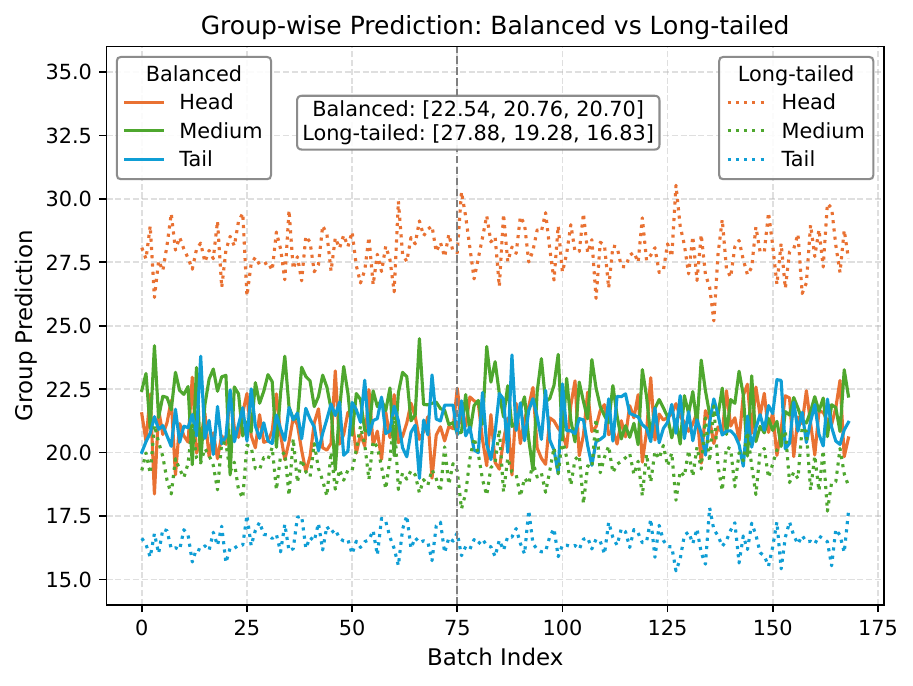}
  \caption{Group-wise prediction trends of the teacher model on the CIFAR-100 dataset across training batches. On the balanced version (solid lines), the teacher produces nearly uniform group-wise outputs. In contrast, on the long-tailed version (dotted lines), the teacher assigns higher probabilities to head classes and lower probabilities to tail classes.}
  \label{fig:group_preds_batch}
\end{figure}

\subsection{Rebalanced cross-group loss}
\label{subsec:inter_group_loss}

\begin{figure*}[t!]
  \centering
  \includegraphics[width=\linewidth]{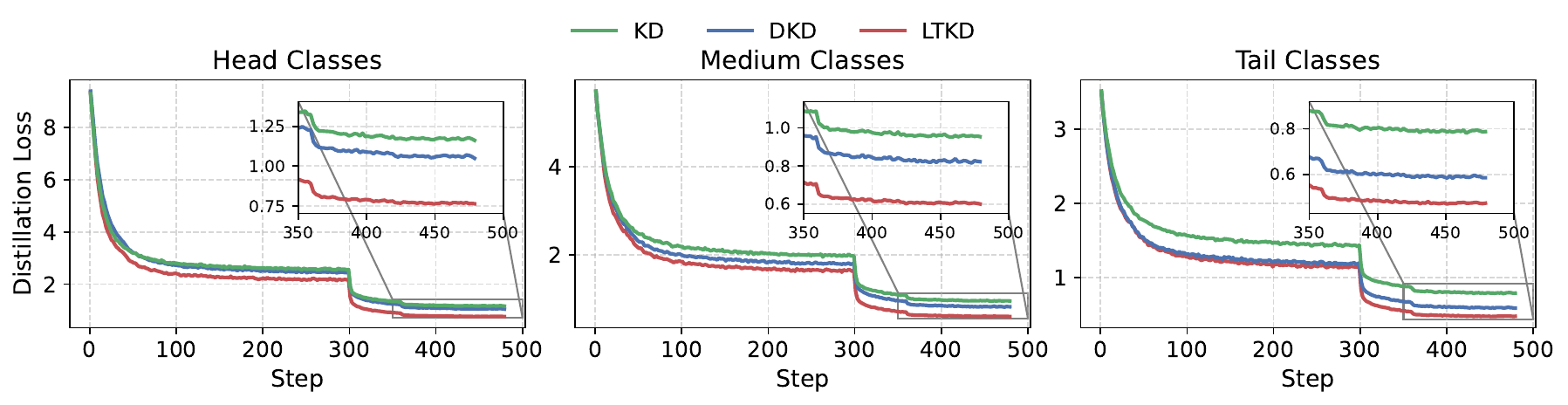}
	\caption{Distillation loss curves for KD, DKD, and LTKD for Head (left), Medium (middle), and Tail (right) class groups. LTKD achieves a consistently lower loss across all three groups compared to the baselines, overcoming the suboptimal convergence caused by teacher bias.}
    \label{fig:loss_curve}
\end{figure*}

In long-tailed classification scenarios, 
a teacher model trained on an imbalanced dataset 
naturally exhibits a prediction bias toward the data-rich head classes. 
To examine this phenomenon, 
we analyze the cross-group output probabilities of a pre-trained teacher model 
$\mathbf{p}^T_\mathcal{G}=[p^T_\mathcal{H}, p^T_\mathcal{M}, p^T_\mathcal{T}]$ 
on both the balanced and long-tailed versions of CIFAR-100.
Specifically, 
we measure the aggregate probability for each class group—head, medium, and tail—
within a single batch. 
On a representative batch from the balanced dataset, 
the teacher model produces nearly uniform group-wise predictions, 
with values of $[22.54, 20.76, 20.70]$. 
In contrast, on the long-tailed dataset, the corresponding values become $[27.88, 19.28, 16.83]$, 
indicating a clear bias toward the head group. 
This trend remains consistent across batches within a single epoch (see \cref{fig:group_preds_batch}), 
indicating that the teacher model systematically assigns higher confidence to head classes.
This imbalance poses a critical challenge in knowledge distillation. 
The KL-based loss incentivizes the student to mimic the teacher's biased outputs, 
thereby forcing it to inherit the same head-class bias. 
Consequently, the student's predictions become skewed toward dominant classes, 
which is detrimental to its tail-class performance.
To mitigate this issue, 
we introduce a group-wise rebalancing mechanism 
that adjusts the teacher’s cross-group probability distribution 
before distillation. 
The goal is 
to align the group-wise probabilities to a uniform distribution—\eg $[21, 21, 21]$—
instead of the skewed values 
observed under long-tailed settings.
This is achieved by calculating scaling factors for each group 
such that their rebalanced predictions become equal across a batch.
Let the sum of cross-group probabilities in a batch $B$ 
be $\mathbf{p}_\text{batch}=[p^B_\mathcal{H}, p^B_\mathcal{M}, p^B_\mathcal{T}]$. 
We define the target average as:
\begin{equation}
\label{eq:target_average}
p^B_\text{avg} = \text{Mean}(p^B_\mathcal{H}, p^B_\mathcal{M}, p^B_\mathcal{T}),
\end{equation}
and set the desired balanced vector as $[p^B_{\text{avg}}, p^B_{\text{avg}}, p^B_{\text{avg}}]$.
The scaling factors for each group are then given by:
\begin{equation}
\label{scaling_factor}
s_\mathcal{H}= \frac{p^B_\text{avg}}{p^B_\mathcal{H}}, s_\mathcal{M} = \frac{p^B_\text{avg}}{p^B_\mathcal{M}}, s_\mathcal{T} = \frac{p^B_\text{avg}}{p^B_\mathcal{T}}.
\end{equation}
However, simply applying these scaling factors for each sample, $\hat{\mathbf{p}}^T_\mathcal{G}=\left[ s_\mathcal{H} p^T_\mathcal{H}, s_\mathcal{M} p^T_\mathcal{M}, s_\mathcal{T} p^T_\mathcal{T} \right]$, does not guarantee a valid probability distribution, 
as the results will not sum to one ($\sum_\mathcal{G} s_\mathcal{G}p^T_\mathcal{G} \neq 1$). 
Therefore, to ensure proper normalization, we apply the following correction:
\begin{equation}
\label{eq:correction}
\hat{\mathbf{p}}^T_\mathcal{G}=\left[\frac{s_\mathcal{H}p^T_\mathcal{H}}{\sum_\mathcal{G} s_\mathcal{G}p^T_\mathcal{G}}, \frac{s_\mathcal{M}p^T_\mathcal{M}}{\sum_\mathcal{G} s_\mathcal{G}p^T_\mathcal{G}}, \frac{s_\mathcal{T}p^T_\mathcal{T}}{\sum_\mathcal{G} s_\mathcal{G}p^T_\mathcal{G}}  \right],
\end{equation}
%
%
where $\hat{\mathbf{p}}^T_\mathcal{G}$ is the rebalanced cross-group probability vector.
This operation rebalances the group-level contributions 
while ensuring the output remains a valid probability distribution. 
Through this, the student is guided by a rebalanced teacher distribution
that no longer disproportionately emphasizes the head group.

\subsection{Reweighted within-group loss}
\label{subsec:intra_group_loss}
We now focus on the within-group loss component, 
which is a weighted sum of KL divergences within each group:
\begin{equation}
\label{eq:intra_loss}
\begin{aligned}
&\sum_\mathcal{G}p^T_\mathcal{G}\cdot\text{KL}\left( \tilde{\mathbf{p}}^T_\mathcal{G} \| \tilde{\mathbf{p}}^S_\mathcal{G} \right) = p^T_\mathcal{H} \cdot \text{KL}\left( \tilde{\mathbf{p}}^T_\mathcal{H} \| \tilde{\mathbf{p}}^S_\mathcal{H} \right) \\
&+ p^T_\mathcal{M} \cdot \text{KL}\left( \tilde{\mathbf{p}}^T_\mathcal{M} \| \tilde{\mathbf{p}}^S_\mathcal{M} \right) + p^T_\mathcal{T} \cdot \text{KL}\left( \tilde{\mathbf{p}}^T_\mathcal{T} \| \tilde{\mathbf{p}}^S_\mathcal{T} \right),
\end{aligned}
\end{equation}
where $\tilde{\mathbf{p}}^T_\mathcal{G}$ and $\tilde{\mathbf{p}}^S_\mathcal{G}$ 
are the within-group distributions of the teacher and student, respectively, 
and $p^T_\mathcal{G}$ denotes the aggregated probability 
that the teacher assigns to group $\mathcal{G}\in\{\mathcal{H}, \mathcal{M}, \mathcal{T}\}$.
In long-tailed classification, 
due to the imbalanced nature of the training data, 
the teacher model tends to allocate higher probability to head classes 
($p^T_\mathcal{H} > p^T_\mathcal{M} > p^T_\mathcal{T}$).
As a result, 
the within-group KL loss becomes heavily biased toward the head group, 
while the contribution from the tail group is substantially diminished. 
This bias causes the student to receive weaker supervision signals for underrepresented classes,
resulting in suboptimal convergence across the full label space (see \cref{fig:loss_curve}).
To address this issue, we propose a reweighting strategy 
that enforces equal importance across all three groups, 
regardless of the teacher's confidence.
Specifically, 
we replace the teacher-derived weights $p^T_\mathcal{G}$ 
with a uniform constant $\beta$, 
yielding the modified within-group loss:
\begin{equation}
\label{eq:modified_intra_loss}
\begin{aligned}
&\beta \sum_\mathcal{G}\text{KL}\left( \tilde{\mathbf{p}}^T_\mathcal{G} \| \tilde{\mathbf{p}}^S_\mathcal{G} \right) \\ 
&= \beta \left(\text{KL}\left( \tilde{\mathbf{p}}^T_\mathcal{H} \| \tilde{\mathbf{p}}^S_\mathcal{H} \right)
+ \text{KL}\left( \tilde{\mathbf{p}}^T_\mathcal{M} \| \tilde{\mathbf{p}}^S_\mathcal{M} \right) + \text{KL}\left( \tilde{\mathbf{p}}^T_\mathcal{T} \| \tilde{\mathbf{p}}^S_\mathcal{T} \right)\right).
\end{aligned}
\end{equation}
This simple modification ensures that 
each group contributes equally to the within-group distillation loss, 
preventing the overrepresented head group from dominating the gradient flow during distillation.

\subsection{Long-tailed knowledge distillation}
\label{subsec:LTKD} 
By combining these two complementary strategies, 
we propose the Long-Tailed Knowledge Distillation (LTKD):
\begin{equation}
\label{eq:ltkd}
\begin{aligned}
\text{LTKD} &= \alpha \cdot \text{KL} \left( \hat{\mathbf{p}}^T_\mathcal{G} \| \mathbf{p}^S_\mathcal{G}\right) + \beta \cdot \sum_\mathcal{G}\text{KL}\left( \tilde{\mathbf{p}}^T_\mathcal{G} \| \tilde{\mathbf{p}}^S_\mathcal{G} \right).
\end{aligned}
\end{equation}
\begin{itemize}
    \item $\hat{\mathbf{p}}^T_\mathcal{G}$ and $\mathbf{p}^S_\mathcal{G}$ denote the rebalanced cross-group distributions from the teacher and the original distribution from the student, respectively,
    \item $\tilde{\mathbf{p}}^T_\mathcal{G}$ and $\tilde{\mathbf{p}}^S_\mathcal{G}$ represent the within-group normalized distributions within each group $\mathcal{G} \in \{\mathcal{H}, \mathcal{M}, \mathcal{T}\}$,
    \item $\alpha$ and $\beta$ are hyperparameters that balance the contributions of the cross- and within-group distillation terms.
\end{itemize}
See the supplementary material for the pseudo-code.

%% file: sec/4_experiment.tex
\section{Experiment}
\label{sec:experiment}

\begin{table*}[]
\caption{Accuracy (\%) on tail $(\mathcal{T})$ and overall (All) classes for CIFAR-100-LT with homogeneous (Top) and heterogeneous (Bottom) architectures. \textbf{Bold} and \underline{underline} denote the best and second-best results, respectively. $\Delta$ is the gap between them.}
\centering
\resizebox{0.79\textwidth}{!}{%
\begin{tabular}{c|cccccc|cccccc}
\toprule
T-S Pairs             & \multicolumn{6}{c|}{ResNet32$\times$4 $-$ ResNet8$\times$4}                                                 & \multicolumn{6}{c}{VGG13 $-$ VGG8}                                                 \\ \midrule
$\gamma$ & \multicolumn{2}{c}{10} & \multicolumn{2}{c}{20} & \multicolumn{2}{c|}{100} & \multicolumn{2}{c}{10} & \multicolumn{2}{c}{20} & \multicolumn{2}{c}{100} \\
Group            & $\mathcal{T}$          & All       & $\mathcal{T}$          & All       & $\mathcal{T}$           & All        & $\mathcal{T}$          & All       & $\mathcal{T}$          & All       & $\mathcal{T}$         & All        \\ \midrule
Teacher & 50.72 & 64.95 & 39.19 & 58.82 & 15.28 & 45.35 & 45.67 & 60.77 & 36.43 & 55.10 & 14.01 & 43.11 \\
Student & 47.32 & 60.59 & 36.99 & 55.44 & 13.38 & 42.48 & 43.67 & 57.43 & 33.89 & 52.29 & 13.13 & 40.70 \\ \midrule
DKD~\cite{dkd} & 49.86 & 64.55 & 37.87 & 58.78 & 13.25 & \underline{46.11} & {48.00} & {61.84} & 37.65 & {56.68} & 14.42 & {44.22} \\
ReviewKD~\cite{review} & \underline{52.08} & {64.71} & \underline{40.12} & \underline{59.17} & \underline{15.09} & 45.91 & 47.75 & 61.43 & {37.69} & 56.51 & \underline{14.76} & 44.19 \\ 
DIST~\cite{dist} & 50.28 & 63.74 & 38.69 & 58.28 & 13.86 & 45.21 & 45.57 & 60.53 & 34.36 & 54.68 & 12.46 & 42.12 \\
CAT-KD~\cite{catkd} & 49.83 & \underline{64.74} & 37.67 & 58.73 & 12.83 & 45.33 & \underline{48.53} & \underline{62.01} & \underline{37.95} & \underline{56.78} & 14.22 & \underline{44.33}  \\
\midrule
\textbf{LTKD} & \textbf{58.66} & \textbf{66.76} & \textbf{49.70} & \textbf{62.54} & \textbf{27.21} & \textbf{51.08} & \textbf{53.95} & \textbf{63.04} & \textbf{45.77} & \textbf{58.86} & \textbf{23.30} & \textbf{47.66}  \\
$\Delta$ & \textbf{+6.58} & \textbf{+2.02} & \textbf{+9.58} & \textbf{+3.37} & \textbf{+12.12} & \textbf{+4.97} & \textbf{+5.42} & \textbf{+1.03} & \textbf{+7.82} & \textbf{+2.08} & \textbf{+8.54} & \textbf{+3.33} \\

\midrule
\midrule

T-S Pairs             & \multicolumn{6}{c|}{WRN--40--2 $-$ ShuffleNetV1}                                                 & \multicolumn{6}{c}{ResNet50 $-$ MobileNetV2}                                                 \\ \midrule
$\gamma$ & \multicolumn{2}{c}{10} & \multicolumn{2}{c}{20} & \multicolumn{2}{c|}{100} & \multicolumn{2}{c}{10} & \multicolumn{2}{c}{20} & \multicolumn{2}{c}{100} \\
Group            & $\mathcal{T}$          & All       & $\mathcal{T}$          & All       & $\mathcal{T}$           & All        & $\mathcal{T}$          & All       & $\mathcal{T}$          & All       & $\mathcal{T}$         & All        \\ \midrule
Teacher & 49.77 & 63.05 & 39.88 & 58.27 & 14.88 & 44.78 & 49.74 & 63.51 & 37.74 & 56.70 & 14.42 & 42.26 \\
Student & 40.04 & 54.06 & 29.91 & 48.22 & 10.74 & 36.21 & 30.47 & 44.58 & 22.32 & 39.25 & 7.04 & 27.56 \\ \midrule
DKD~\cite{dkd} & 50.86 & 63.65 & 39.94 & 58.28 & 15.04 & 45.24 & \underline{43.29} & {57.20} & \underline{33.23} & \underline{52.20} & \underline{12.45} & \underline{39.21} \\
ReviewKD~\cite{review} & \underline{51.24} & \underline{63.90} & \underline{40.44} & \underline{58.63} & \underline{15.81} & \underline{45.40} & 33.68 & 47.75 & 24.80 & 42.08 & 9.75 & 31.86 \\ 
DIST~\cite{dist} & 48.40 & 62.47 & 37.48 & 56.92 & 12.23 & 41.95 & 37.86 & 52.36 & 27.11 & 46.50 & 9.81 & 34.96 \\
CAT-KD~\cite{catkd} & 51.02 & 63.68 & 40.23 & 58.26 & 14.68 & 44.84 & 43.18 & \underline{57.23} & 33.17 & 51.90 & 11.61 & 38.45 \\

\midrule
\textbf{LTKD} & \textbf{57.40} & \textbf{65.42} & \textbf{48.42} & \textbf{60.94} & \textbf{23.99} & \textbf{48.60} & \textbf{48.43} & \textbf{57.79} & \textbf{40.82} & \textbf{53.70} & \textbf{21.04} & \textbf{42.45} \\
$\Delta$ & \textbf{+6.16} & \textbf{+1.52} & \textbf{+7.98} & \textbf{+2.31} & \textbf{+8.18} & \textbf{+3.20} & \textbf{+5.14} & \textbf{+0.56} & \textbf{+7.59} & \textbf{+1.50} & \textbf{+8.59} & \textbf{+3.24} \\
\bottomrule

\end{tabular}%
}
\label{table:CIFAR-100_lt}
\end{table*}

\subsection{Dataset}
\label{subsec:dataset}
We construct CIFAR-100-LT, TinyImageNet-LT, and ImageNet-LT by 
introducing class imbalance 
into the balanced CIFAR-100~\cite{cifar}, TinyImageNet~\cite{tiny}, and ImageNet~\cite{imagenet} datasets, 
following previous works~\cite{cifarlt, imagenetlt, globallt}.
For each class $c$, the number of training samples, 
denoted by $|\mathcal{D}_c|$, 
is reduced using an exponential decay function:
\begin{equation}
\label{eq:decay_func}
|\hat{\mathcal{D}}_c|=|\mathcal{D}_c|\cdot\gamma^{-c/C},
\end{equation}
where $C$ is the total number of classes 
and $\gamma$ is the imbalance factor, 
defined as the ratio between the number of training samples 
in the largest and smallest classes~\cite{cifarlt}.
For both CIFAR-100-LT and TinyImageNet-LT, 
we construct three versions of the training set with imbalance factors of $[10, 20, 100]$. 
For ImageNet-LT, imbalance factors of $[5, 10, 20]$ are used.
The classes are divided into three groups 
based on the number of training samples: 
Head (the top 33\% of classes), Medium (the next 34\%), and Tail (the bottom 33\%).
The test set remains balanced to ensure fair evaluation.

\subsection{Implementation details}
\label{subsec:implement}
We conduct experiments using widely used CNN architectures, 
including ResNet~\cite{resnet}, VGG~\cite{vgg}, WideResNet (WRN)~\cite{wideresnet}, ShuffleNet~\cite{shufflenet, shufflenetv2}, and MobileNet~\cite{mobilenet}. 
Each experiment is conducted three times, and the average performance is reported. 
Additional details and results are provided in the supplementary material.

\subsection{Main results}
\label{subsec:main_results}
\paragraph{CIFAR-100-LT.}
In all settings, 
LTKD consistently shows remarkable improvements 
in both overall accuracy (All) and tail-class accuracy ($\mathcal{T}$), 
as shown in~\cref{table:CIFAR-100_lt}. 
For instance, with a ResNet32$\times$4$-$ResNet8$\times$4 pair ($\gamma=100$), 
LTKD boosts the tail-class accuracy from 15.09\% to 27.21\%, 
and overall accuracy from 46.11\% to 51.08\%. 
Similarly, for the ResNet50$–$MobileNetV2 ($\gamma=100$) pair, 
LTKD delivers a substantial gain of +8.59\% in tail accuracy and +3.24\% in overall accuracy. 
These gains are consistent across all imbalance levels and architecture combinations, 
underscoring the importance of our rebalanced and reweighted supervision.

\begin{table*}[th]
\caption{Accuracy (\%) on tail $(\mathcal{T})$ and overall (All) classes for TinyImageNet-LT with homogeneous (Top) and heterogeneous (Bottom) architectures. \textbf{Bold} and \underline{underline} denote the best and second-best results, respectively. $\Delta$ is the gap between them.}
\centering
\resizebox{0.79\textwidth}{!}{%
\begin{tabular}{c|cccccc|cccccc}
\toprule
T-S Pairs             & \multicolumn{6}{c|}{ResNet32$\times$4 $-$ ResNet8$\times$4}                                                 & \multicolumn{6}{c}{VGG13 $-$ VGG8}                                                 \\ \midrule
$\gamma$ & \multicolumn{2}{c}{10} & \multicolumn{2}{c}{20} & \multicolumn{2}{c|}{100} & \multicolumn{2}{c}{10} & \multicolumn{2}{c}{20} & \multicolumn{2}{c}{100} \\
Group            & $\mathcal{T}$          & All       & $\mathcal{T}$          & All       & $\mathcal{T}$           & All        & $\mathcal{T}$          & All       & $\mathcal{T}$          & All       & $\mathcal{T}$         & All        \\ \midrule
Teacher & 38.47 & 52.64 & 28.74 & 47.49 & 9.53 & 35.37 & 32.53 & 45.23 & 21.85 & 39.75 & 6.29 & 29.71 \\
Student & 29.72 & 44.60 & 21.46 & 40.25 & 4.73 & 30.62 & 31.12 & 43.76 & 22.19 & 39.00 & 6.88 & 29.96 \\ \midrule
KD~\cite{hinton} & 27.34 & 45.38 & 18.11 & 41.35 & 3.38 & 31.42 & 31.92 & 47.16 & 20.60 & 41.49 & 3.99 & 30.95 \\
DKD~\cite{dkd} & {34.70} & 48.93 & \underline{26.58} & {44.84} & \underline{9.09} & \underline{34.61} & 33.20 & \underline{48.01} & 22.65 & \underline{42.44} & 5.88 & 31.82  \\
ReviewKD~\cite{review} & 32.85 & {49.13} & 23.43 & 44.62 & 5.39 & 33.51 & \underline{34.39} & 47.66 & \underline{24.84} & 42.36 & \underline{7.61} & \underline{32.18} \\ 
DIST~\cite{dist} & \underline{34.81} & \underline{50.14} & 25.71 & \underline{45.52} & 7.30 & 33.98 & 33.48 & 47.22 & 23.19 & 41.46 & 5.98 & 31.01 \\

\midrule
\textbf{LTKD} & \textbf{40.66} & \textbf{51.33} & \textbf{31.33} & \textbf{47.05} & \textbf{10.48} & \textbf{36.21} & \textbf{38.90} & \textbf{49.43} & \textbf{29.30} & \textbf{44.22} & \textbf{9.73} & \textbf{33.78} \\
$\Delta$ & \textbf{+5.85} & \textbf{+1.19} & \textbf{+4.75} & \textbf{+1.53} & \textbf{+1.39} & \textbf{+1.60} & \textbf{+4.51} & \textbf{+1.42} & \textbf{+4.46} & \textbf{+1.78} & \textbf{+2.12} & \textbf{+1.60} \\

\midrule
\midrule

T-S Pairs             & \multicolumn{6}{c|}{ResNet32$\times$4 $-$ ShuffleNetV1}                                                 & \multicolumn{6}{c}{VGG13 $-$ MobileNetV2}                                                 \\ \midrule
$\gamma$ & \multicolumn{2}{c}{10} & \multicolumn{2}{c}{20} & \multicolumn{2}{c|}{100} & \multicolumn{2}{c}{10} & \multicolumn{2}{c}{20} & \multicolumn{2}{c}{100} \\
Group            & $\mathcal{T}$          & All       & $\mathcal{T}$          & All       & $\mathcal{T}$           & All        & $\mathcal{T}$          & All       & $\mathcal{T}$          & All       & $\mathcal{T}$         & All        \\ \midrule
Teacher & 38.47 & 52.64 & 28.74 & 47.49 & 9.53 & 35.37 & 32.53 & 45.23 & 21.85 & 39.75 & 6.29 & 29.71 \\
Student & 24.43 & 37.10 & 16.80 & 31.85 & 4.71 & 22.77 & 26.98 & 39.73 & 17.38 & 33.14 & 3.97 & 22.99 \\ \midrule
KD~\cite{hinton} & 34.67 & 49.05 & 24.12 & 42.81 & 5.62 & 30.74 & 31.24 & 45.60 & 19.53 & 39.50 & 3.27 & 28.44 \\
DKD~\cite{dkd} & \underline{36.83} & \underline{50.22} & 26.64 & 44.38 & 8.34 & \underline{33.23} & \underline{33.07} & \underline{46.79} & 22.06 & \underline{40.87} & 5.70 & \underline{29.97} \\
ReviewKD~\cite{review} & 36.30 & 49.27 & \underline{27.09} & \underline{44.72} & \underline{8.49} & 33.12 & 32.37 & 44.99 & \underline{22.77} & 39.36 & \underline{7.02} & 29.20 \\ 
DIST~\cite{dist} & 36.49 & 50.07 & 25.74 & 43.59 & 7.23 & 31.19 & 32.92 & 46.09 & 21.77 & 40.05 & 5.52 & 29.08 \\

\midrule
\textbf{LTKD} & \textbf{42.12} & \textbf{51.64} & \textbf{33.06} & \textbf{46.41} & \textbf{12.85} & \textbf{35.09} & \textbf{39.04} & \textbf{48.71} & \textbf{28.28} & \textbf{43.22} & \textbf{9.52} & \textbf{32.30} \\
$\Delta$ & \textbf{+5.29} & \textbf{+1.42} & \textbf{+5.97} & \textbf{+1.69} & \textbf{+4.36} & \textbf{+1.86} & \textbf{+5.97} & \textbf{+1.92} & \textbf{+5.51} & \textbf{+2.35} & \textbf{+2.50} & \textbf{+2.33} \\
\bottomrule

\end{tabular}%
} 
\label{table:tinyimagenet_lt}
\end{table*}

\paragraph{TinyImageNet-LT.}
%
%
Even on the more challenging TinyImageNet-LT, 
LTKD proves its effectiveness with consistent performance improvements, 
as shown in~\cref{table:tinyimagenet_lt}. 
For example, with the ResNet32$\times$4$–$ResNet8$\times$4 pair ($\gamma=100$), 
LTKD improves tail accuracy from 9.09\% to 10.48\% and overall accuracy from 34.61\% to 36.21\%. 
Similarly, for VGG13$–$MobileNetV2 pair ($\gamma=100$), 
LTKD boosts tail-class accuracy from 7.02\% to 9.52\% and overall accuracy from 29.97\% to 32.30\%. 
These results highlight the robustness of LTKD on a more complex benchmark.

\begin{table*}[]
\caption{Accuracy (\%) on tail $(\mathcal{T})$ and overall (All) classes for ImageNet-LT. \textbf{Bold} and \underline{underline} denote the best and second-best results, respectively. $\Delta$ is the gap between them.
}
\centering
\resizebox{0.79\textwidth}{!}{%
\begin{tabular}{c|cccccc|cccccc}
\toprule
T-S Pairs             & \multicolumn{6}{c|}{ResNet34 $-$ ResNet18}                                                 & \multicolumn{6}{c}{ResNet50 $-$ MobileNetV1}                                                 \\ \midrule
$\gamma$ & \multicolumn{2}{c}{5} & \multicolumn{2}{c}{10} & \multicolumn{2}{c|}{20} & \multicolumn{2}{c}{5} & \multicolumn{2}{c}{10} & \multicolumn{2}{c}{20} \\
Group            & $\mathcal{T}$          & All       & $\mathcal{T}$          & All       & $\mathcal{T}$           & All        & $\mathcal{T}$          & All       & $\mathcal{T}$          & All       & $\mathcal{T}$         & All        \\ \midrule
Teacher & 57.75 & 67.61 & 50.61 & 63.85 & 43.00 & 59.91 & 59.62 & 69.06 & 52.69 & 65.57 & 44.82 & 61.46 \\
Student & 54.48 & 64.73 & 47.45 & 61.18 & 39.98 & 57.25 & 55.42 & 65.46 & 49.49 & 62.55 & 41.68 & 58.94 \\ \midrule
KD~\cite{hinton} & 56.14 & 66.27 & 49.35 & 63.03 & 41.72 & 59.15 & 56.58 & 66.51 & 50.42 & 63.70 & 42.45 & 59.91 \\
DKD~\cite{dkd} & 56.83 & 66.84 & 49.95 & 63.50 & 42.65 & 59.82 & 58.54 & 68.09 & 52.39 & 65.04 & 45.04 & 61.46 \\
ReviewKD~\cite{review} & \underline{57.27} & \underline{66.96} & \underline{50.80} & \underline{63.72} & \underline{43.48} & \underline{60.15} & \underline{58.59} & \underline{68.10} & \underline{53.06} & \underline{65.31} & \underline{45.35} & \underline{61.84} \\ 
DIST~\cite{dist} & 56.79 & 66.66 & 50.28 & 63.56 & 42.78 & 59.94 & 57.29 & 66.94 & 50.89 & 64.09 & 43.70 & 60.71 \\
CAT-KD~\cite{catkd} & 55.92 & 66.14 & 49.58 & 63.20 & 42.20 & 59.67 & 57.08 & 66.96 & 51.00 & 64.07 & 43.55 & 60.54 \\

\midrule
\textbf{LTKD} & \textbf{58.33} & \textbf{67.23} & \textbf{52.55} & \textbf{64.29} & \textbf{45.88} & \textbf{60.80} & \textbf{60.17} & \textbf{68.48} & \textbf{54.40} & \textbf{65.52} & \textbf{48.55} & \textbf{62.22} \\
$\Delta$ & \textbf{+1.06} & \textbf{+0.27} & \textbf{+1.75} & \textbf{+0.57} & \textbf{+2.40} & \textbf{+0.65} & \textbf{+1.58} & \textbf{+0.38} & \textbf{+1.34} & \textbf{+0.21} & \textbf{+3.20} & \textbf{+0.38} \\
\bottomrule
\end{tabular}%
} 
\label{table:imagenet_lt}
\end{table*}

\begin{table*}[]
\caption{Performance comparison of cross-group loss on CIFAR-100-LT ($\gamma=20$). ``Biased" refers to the unbalanced cross-group loss, while ``Ours" applies our balanced loss. The within-group loss is disabled for both methods. $\mathcal{T}$ and All denote tail and overall accuracy, respectively. $\Delta$ shows the improvement of ``Ours" over the ``Biased" baseline.}
\centering
\resizebox{0.8\textwidth}{!}{%
\begin{tabular}{ccc|cccccccccccc}
\toprule
\multirow{2}{*}{Models} & \multicolumn{2}{c|}{Teacher} & \multicolumn{4}{c}{ResNet32$\times$4}                                   & \multicolumn{4}{c}{VGG13}                                  & \multicolumn{4}{c}{WRN--40--2}                                    \\
                        & \multicolumn{2}{c|}{Student} & \multicolumn{2}{c}{ResNet8$\times$4} & \multicolumn{2}{c}{ShuffleNetV1} & \multicolumn{2}{c}{VGG8} & \multicolumn{2}{c}{MobileNetV2} & \multicolumn{2}{c}{WRN--40--1} & \multicolumn{2}{c}{ShuffleNetV1} \\ \midrule
Loss                    & Cross        & Within        & $\mathcal{T}$             & All           & $\mathcal{T}$              & All             & $\mathcal{T}$          & All         & $\mathcal{T}$             & All             & $\mathcal{T}$             & All            & $\mathcal{T}$              & All             \\ \midrule
Biased & \cmark & \xmark & 38.30 & 55.77 & 30.01 & \textbf{48.82} & 35.55 & 53.30 & 23.42 & 40.29 & 34.64 & 53.97 & 30.23 & 49.31 \\
\textbf{Ours} & \cmark & \xmark & \textbf{40.51} & \textbf{56.55} & \textbf{30.68} & {48.81} & \textbf{37.26} & \textbf{53.70} & \textbf{23.97} & \textbf{40.30} & \textbf{35.23} & \textbf{54.35} & \textbf{32.44} & \textbf{49.91} \\ \midrule
\multicolumn{3}{c}{$\Delta$} & \textbf{+2.21} & \textbf{+0.78} & \textbf{+0.67} & -0.01 & \textbf{+1.71} & \textbf{+0.40} & \textbf{+0.55} & \textbf{+0.01} & \textbf{+0.59} & \textbf{+0.38} & \textbf{+2.21} & \textbf{+0.60} \\  \bottomrule
\end{tabular}}
\label{table:inter_loss_ablation}
\end{table*}

\begin{table*}[]
\caption{Ablation study of cross-group and within-group losses on CIFAR-100-LT ($\gamma=20$). ``Baseline" is vanilla KD. ``Ours (Cross \cmark, Within \cmark)" combines both rebalanced cross-group and reweighted within-group loss.}
\centering
\resizebox{0.8\textwidth}{!}{%
\begin{tabular}{ccc|cccccccccccc}
\toprule
\multirow{2}{*}{Models} & \multicolumn{2}{c|}{Teacher} & \multicolumn{4}{c}{ResNet32$\times$4}                                   & \multicolumn{4}{c}{VGG13}                                  & \multicolumn{4}{c}{WRN--40--2}                                    \\
                        & \multicolumn{2}{c|}{Student} & \multicolumn{2}{c}{ResNet8$\times$4} & \multicolumn{2}{c}{ShuffleNetV1} & \multicolumn{2}{c}{VGG8} & \multicolumn{2}{c}{MobileNetV2} & \multicolumn{2}{c}{WRN--40--1} & \multicolumn{2}{c}{ShuffleNetV1} \\ \midrule
Loss                    & Cross        & Within        & $\mathcal{T}$             & All           & $\mathcal{T}$              & All             & $\mathcal{T}$          & All         & $\mathcal{T}$             & All             & $\mathcal{T}$             & All            & $\mathcal{T}$              & All             \\ \midrule
Baseline & \xmark & \xmark & 36.81 & 57.41 & 36.13 & 55.03 & 37.29 & 56.13 & 29.05 & 47.69 & 35.63 & 56.71 & 38.30 & 56.61 \\
\multirow{3}{*}{\textbf{Ours}} & \cmark & \xmark & 40.51 & 56.55 & 30.68 & 48.81 & 37.26 & 53.70 & 23.97 & 40.30 & 35.23 & 54.35 & 32.44 & 49.91 \\
& \xmark & \cmark & 42.34 & 59.78 & 39.10 & 56.84 & 38.54 & 56.83 & 31.99 & 49.20 & 39.67 & 58.11 & 40.45 & 57.65 \\
& \cmark & \cmark & \textbf{49.70} & \textbf{62.54} & \textbf{45.94} & \textbf{59.62} & \textbf{45.77} & \textbf{58.86} & \textbf{38.33} & \textbf{52.03} & \textbf{45.74} & \textbf{59.91} & \textbf{48.42} & \textbf{60.94} \\ \bottomrule
\end{tabular}
} 
\label{table:each_components}
\end{table*}

\vspace{-11pt}
\paragraph{ImageNet-LT.}
On the large-scale ImageNet-LT, 
LTKD demonstrates its scalability 
by consistently outperforming all baselines. 
For instance, with a ResNet34$–$ResNet18 pair, 
LTKD improves tail accuracy by +1.06\%, +1.75\%, and +2.40\% for $\gamma=5,10,20$, respectively. 
This strong performance extends to the ResNet50$–$MobileNetV1 setting, 
with gains of up to +3.20\% on tail classes, 
highlighting the method's ability to perform effective knowledge transfer 
even on large-scale, severely imbalanced datasets.

\subsection{Ablation study}
\label{subsec:ablation}
\paragraph{Rebalance effect.}
To isolate the effect of our rebalanced cross-group loss, 
we compare it against a non-rebalanced baseline, 
disabling the within-group loss for both settings (\cref{table:inter_loss_ablation}). 
Our rebalanced loss consistently improves tail accuracy, 
improving it from 38.30\% to 40.51\% for the ResNet32$\times$4$–$ResNet8$\times$4 pair. 
Similar positive trends are observed in other architecture pairs, 
confirming the effectiveness of the rebalance process, 
particularly for underrepresented classes. 
Interestingly, we observed a minor trade-off in the ResNet32$\times$4$–$ShuffleNetV1 pair, 
where tail accuracy improved (30.01\% to 30.68\%) 
but overall accuracy slightly dropped (-0.01\%), 
likely due to suppressing head-class signals 
during the rebalance process. 
However, this is resolved by incorporating the within-group loss (\cref{table:each_components}),
demonstrating the complementary strength of our LTKD.

\vspace{-11pt}

\paragraph{Impact of each component.}
We conduct an ablation study 
to disentangle the contributions of our cross-group and within-group components (\cref{table:each_components}). 
Our findings show that 
each component independently boosts both tail and overall performance.
For the ResNet32$\times$4$-$ResNet8$\times$4 pair, 
the cross-group loss alone yields a +3.70\% improvement in tail accuracy, 
while the within-group loss alone provides an even larger gain of +5.53\%. 
This highlights that uniformly weighting within-group losses 
is particularly effective in mitigating teacher bias.
The best performance is consistently achieved 
when both strategies are combined. 
In the VGG13$-$MobileNetV2 setting, 
our LTKD improves tail accuracy by +9.28\% and overall accuracy by +4.34\% over the baseline. 
These results confirm that 
the two components are complementary, 
and jointly addressing cross- and within-group biases is key 
to effective long-tailed knowledge distillation.

\begin{figure}[]
  \centering
  \includegraphics[width=\linewidth]{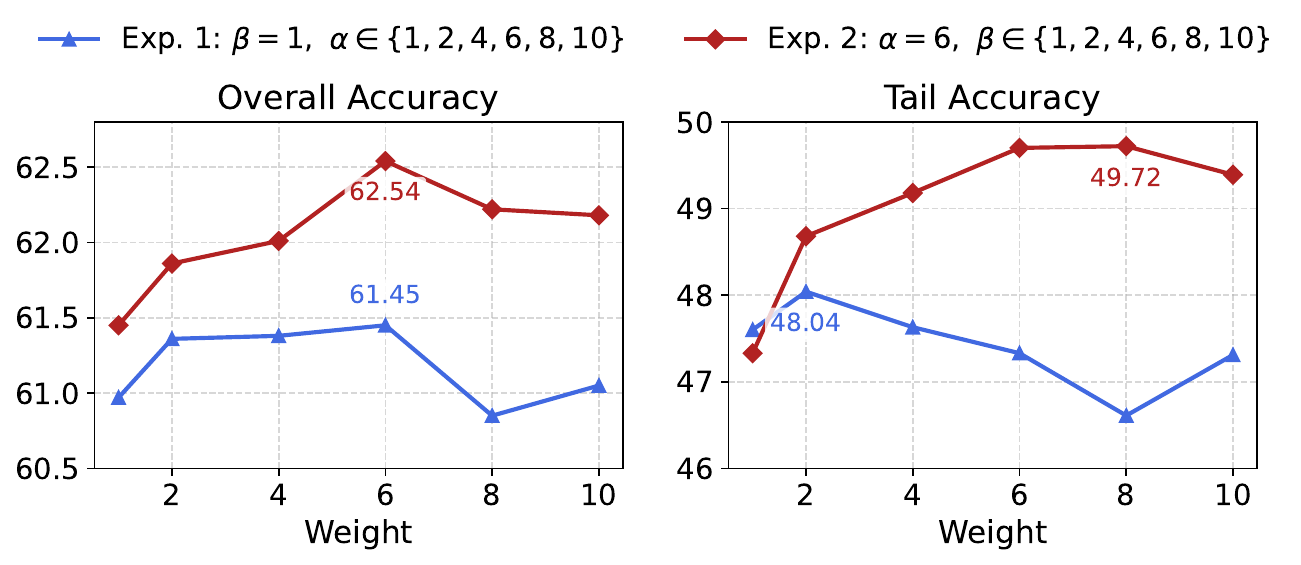}
	\caption{
    Hyperparameter sensitivity of LTKD on CIFAR-100-LT ($\gamma=20$), showing overall accuracy (Left) and tail-class accuracy (Right). The blue line shows the effect of varying the cross-group weight $\alpha$ (while $\beta=1$). The red line shows the effect of varying the within-group weight $\beta$ (while $\alpha=6$).}
    \label{fig:ltkd_ablation_hyper}
\end{figure}

\paragraph{Hyperparameters.}
We analyze the sensitivity 
to hyperparameters $\alpha$ (cross-group) and $\beta$ (within-group) on CIFAR-100-LT ($\gamma=20$) 
using a ResNet32$\times$4 and ResNet8$\times$4 pair (\cref{fig:ltkd_ablation_hyper}). 
Varying $\alpha$ (with $\beta=1$) shows overall accuracy peaking at $\alpha=6$ (61.45\%). 
When fixing $\alpha=6$ and varying $\beta$, performance peaks at $\beta=6$ (62.54\%). 
Tail accuracy follows a similar trend, reaching its peak of 49.72\% ($\alpha=6, \beta=8$). 
Importantly, LTKD consistently outperforms 
the strong ReviewKD baseline (Overall: 59.17\%, Tail: 40.12\%) 
across a wide range of $\alpha$ and $\beta$ values, 
demonstrating both its robustness and its significant benefit for underrepresented classes. 

\begin{table}[]
\caption{Overall accuracy on CIFAR-100-LT $(\gamma=100)$ with varying number of groups, including the continuous reweighting.}
\label{tab:groups}
\resizebox{\columnwidth}{!}{%
\begin{tabular}{c|cccccccc}
\toprule
$n(\mathcal{G})$ & 3     & 4     & 5     & 10    & 20    & 25    & 50    & 100   \\
\midrule
R32$\times$4$-$R8$\times$4 & 51.08 & 51.08 & \underline{51.10} & \textbf{51.14} & 50.99 & 50.34 & 50.06 & 50.41 \\
VGG13$-$VGG8   & 47.66 & 47.85 & 48.06 & 48.26 & \textbf{48.69} & \underline{48.58} & 48.19 & 47.82
\\
WRN402$-$SV1   & 48.60 & 48.98 & 49.03 & 49.27 & 49.54 & \textbf{49.90} & \underline{49.64} & 47.95 \\
R50$-$MV2    & 42.45 & 43.01 & 42.99 & 43.40 & \textbf{43.43} & \underline{43.42} & 42.51 & 41.18 \\
\bottomrule
\end{tabular}%
}
\end{table}

\vspace{-12pt}

\paragraph{Number of groups.}

To evaluate sensitivity to group definitions, we vary the number of groups $n(\mathcal{G})$ from 3 to 100 (Table~\ref{tab:groups}). Even at the simple $n(\mathcal{G})=3$ $(\mathcal{H}, \mathcal{M}, \mathcal{T})$, LTKD already achieves significant improvements over previous methods, and increasing $n(\mathcal{G})$ enables more fine-grained bias correction, yielding consistent performance gains.

The $n(\mathcal{G})=100$ setting corresponds to a continuous reweighting scheme, where LTKD continues to outperform existing methods (see Table~\ref{table:CIFAR-100_lt}), demonstrating that the framework naturally extends beyond discrete grouping. Nevertheless, the discrete grouping design enables the explicit formulation of the within-group loss, leading to additional performance gains.

%% file: sec/5_conclusion.tex
\section{Conclusion}
\label{sec:conclusion}
We present Long-Tailed Knowledge Distillation (LTKD), 
a novel framework 
addressing the ineffective transfer of biased knowledge from teachers trained on long-tailed datasets. 
LTKD introduces a rebalanced cross-group loss and a reweighted within-group loss, 
designed based on our decomposition of the conventional KL divergence, 
which revealed its inherent bias. 
This enables the effective distillation of balanced knowledge from a biased teacher. 
Through extensive experiments, 
we demonstrate that LTKD consistently enhances both overall and tail-class performance 
in long-tailed scenarios. 
We plan to extend this framework to other domains where long-tail issues are pervasive, 
such as object detection and semantic segmentation.